%% file: emnlp2018.tex
\documentclass[11pt,a4paper]{article}
\usepackage[hyperref]{emnlp2018}
\usepackage{times}
\usepackage{natbib}
\usepackage{latexsym}
\usepackage{comment}
\usepackage{url}
\usepackage{graphicx}
\usepackage{amsmath}
\usepackage{amssymb}
\usepackage{multirow}
\usepackage{algorithm}
\usepackage{algorithmic}
\urldef{\mailsc}\path|{svmehta, jaylee, jgc}@cs.cmu.edu| 
\aclfinalcopy 

\newcommand\BibTeX{B{\sc ib}\TeX}
\newcommand\confname{EMNLP 2018}
\newcommand\conforg{SIGDAT}
\newcommand{\cut}[1]{}
\newcommand*{\x}{\ensuremath{\mathbf{x}}}
\newcommand*{\y}{\ensuremath{\mathbf{y}}}
\newcommand*{\w}{\ensuremath{\mathbf{w}}}
\newcommand*{\f}{\ensuremath{\mathbf{f}}}

\DeclareMathOperator*{\argmax}{arg\,max}
\newcommand*{\jy}[1]{#1}
\newcommand*{\tosvm}[1]{}
\newcommand*{\svm}[1]{#1}

\title{Towards Semi-Supervised Learning for Deep Semantic Role Labeling}

\author{Sanket Vaibhav Mehta\thanks{\hspace{5mm} Equal contribution, name order decided by coin flip.}
\And
Jay Yoon Lee\footnotemark[1] \\
 \\
School of Computer Science, \\
Carnegie Mellon University, \\
Pittsburgh, PA \\ 
\path|{svmehta, jaylee, jgc}@cs.cmu.edu|\\
\And Jaime Carbonell \\
}

\date{}

\begin{document}
\maketitle
\begin{abstract}
  \input{0.abstract.tex}
\end{abstract}

\input{1.introduction.tex}

\input{2.methodology.tex}


\input{3.experiments.tex}

\input{4.related_work.tex}

\input{5.conclusion.tex}
\bibliography{emnlp2018}
\bibliographystyle{acl_natbib_nourl}

\newpage
\section*{Appendix}
\appendix
\input{appendix}

\end{document}

%% file: 0.abstract.tex

Neural models have shown several state-of-the-art performances on Semantic Role Labeling (SRL). However, the neural models require an immense amount of semantic-role corpora and are thus not well suited for low-resource languages or domains. 
The paper proposes a semi-supervised semantic role labeling method that outperforms the state-of-the-art in limited SRL training corpora. The method is based on explicitly enforcing syntactic constraints by augmenting the training objective with a syntactic-inconsistency loss component and uses SRL-unlabeled instances to train a joint-objective LSTM. 
\jy{On CoNLL-2012 English section, the proposed semi-supervised training with 1\%, 10\% SRL-labeled data and varying amounts of SRL-unlabeled data achieves +1.58, +0.78 F1, respectively, over the pre-trained models that were trained on SOTA architecture with ELMo on the same SRL-labeled data.}
Additionally, by using the syntactic-inconsistency loss on inference time, the proposed model achieves +3.67, +2.1 F1 over pre-trained model on 1\%, 10\% SRL-labeled data, respectively.

%% file: 1.introduction.tex

\section{Introduction}
\label{intro}
Semantic role labeling (SRL), a.k.a shallow semantic parsing, identifies the arguments corresponding to each clause or proposition, i.e. its semantic roles, based on lexical and positional information. SRL labels non-overlapping text spans corresponding to typical semantic roles such as Agent, Patient, Instrument, Beneficiary, etc. This task finds its use in many downstream applications such as question-answering \cite{shen2007using}, information extraction \cite{bastianelli2013textual}, machine translation, etc.

Several SRL systems relying on large annotated corpora have been proposed \cite{peters2018deep, he2017deep}, and perform relatively well. A more challenging task is to design an SRL method for low resource scenarios (e.g. rare languages or domains) where we have limited annotated data but where we may leverage annotated data from related tasks. Therefore, in this paper, we focus on building effective systems for low resource scenarios and illustrate our system's performance by simulating low resource scenarios for English. 

SRL systems for English are built using large annotated corpora of verb predicates and their arguments provided as part of the PropBank and OntoNotes v5.0 projects \cite{kingsbury2002treebank, pradhan2013towards}. These corpora are built by adding semantic role annotations to the constituents of previously-annotated syntactic parse trees in the Penn Treebank \cite{marcus1993building}. Traditionally, SRL relies heavily on using syntactic parse trees either from shallow syntactic parsers (chunkers) or full syntactic parsers and \citeauthor{punyakanok2008importance} shows significant improvements by using syntactic parse trees.
\cut{Further, work by \cite{punyakanok2008importance} focus on quantifying the importance of syntactic parsing and inference for SRL and show that syntactic parse trees does help in improving SRL systems. }

Recent breakthroughs motivated by end-to-end deep learning techniques \cite{zhou2015end, he2017deep} achieve state-of-the-art performance without leveraging any syntactic signals, relying instead on ample role-label annotations. 
We hypothesize that by leveraging syntactic structure while training neural SRL models,
we may achieve robust performance, especially for low resource scenarios. Specifically, we propose to leverage syntactic parse trees as hard constraints for the SRL task i.e., we explicitly enforce that the predicted argument spans of the SRL network must agree with the spans implied by the syntactic parse of the sentence 
via scoring function in the training objective.
Moreover, we present a semi-supervised learning (SSL) based formulation, wherein we leverage syntactic parse trees for SRL-unlabeled data to build effective SRL for low resource scenarios. 

We build upon the state-of-the-art SRL system by \cite{peters2018deep, he2017deep}, where we formulate SRL as a BIO tagging problem and use multi-layer highway bi-directional LSTMs. However, we differ in terms of our training objective. In addition to the log-likelihood objective, we also include syntactic inconsistency loss (defined in Section \ref{sec:synt-const}) which quantifies the hard constraint (spans implied by syntactic parse) violations in our training objective. In other words, while training our model, we enforce the outputs of our system to agree with the spans implied by the syntactic parse of the sentence as much as possible. In summary, our contributions to low-resource SRL are:
\begin{enumerate}
\item A novel formulation which leverages syntactic parse trees for SRL by introducing them as hard constraints while training the model. 
\item Experiments with varying amounts of SRL-unlabeled data that point towards semi-supervised learning for low-resource SRL by leveraging the fact that syntactic inconsistency loss does not require labels.
\end{enumerate} 

%% file: 2.methodology.tex

\section{Proposed Approach}
We build upon an existing deep-learning approach to SRL \cite{he2017deep}. First we revisit definitions introduced by \cite{he2017deep} and then discuss about our formulation. 
\subsection{Task definition}
Given a sentence-predicate pair $(\x, v)$, SRL task is defined as predicting a sequence of tags $\y$, where each $y_i$ belongs to a set of BIO tags ($\Omega$). So, for an argument span with semantic role $\text{ARG}_{i}$, $\text{B-ARG}_{i}$ tag indicates that the corresponding token marks the beginning of the argument span and $\text{I-ARG}_{i}$ tag indicates that the corresponding token is inside of the argument span and O tag indicates that token is outside of all argument spans. Let $n = |\x| = |\y|$ be the length of the sentence. Further, let \textit{srl-spans}$(\y)$ denote the set of all argument spans in the SRL tagging $\y$. Similarly, \textit{parse-spans}$(\x)$ denotes the set of all unlabeled parse constituents for the given sentence $\x$. \svm{Lastly, SRL-labeled/unlabeled data refers to sentence-predicate pairs with/without gold SRL tags.}
\subsection{State-of-the-Art (SOTA) Model}
\label{sec:sota}

\citeauthor{he2017deep} proposed a deep bi-directional LSTM to learn a locally decomposed scoring function conditioned on the entire input sentence- $\sum_{i=1}^{n}\log p(y_i | \x)$. 
To learn the parameters of a network, the conditional negative log-likelihood $\mathcal{L}(\w)$ of a sample of training data $\mathcal{T} = \lbrace \x^{(t)}, \y^{(t)}\rbrace_{\forall t}$ is minimized, where $\mathcal{L}(\w)$ is 
\begin{align}
\mathcal{L}(\w) &= -\sum_{(\x,\y) \in \mathcal{T}} \sum_{i=1}^{|\y|} \log p(y_i|\x;\w). \label{eq:nll-loss}
\end{align}

Since Eq.(\ref{eq:nll-loss}) does not model any dependencies between the output tags, the predicted output tags tend to be structurally inconsistent. 
To alleviate this problem, \cite{he2017deep} searches for the best $\hat{\y}$
over the space of all possibilities ($\Omega^n$)
using the scoring function $f(x,y)$, which incorporates log probability and structural penalty terms. The details of scoring function is on Appendix Eq.(\ref{eq:score_function}).
\begin{align}
\hat{\y} = \argmax_{\y' \in \Omega^n} \f(\x, \y') \label{eq:argmax_y}
\end{align}

\subsection{Structural Constraints}

\svm{
There are different types of structural constraints: BIO, SRL and syntactic constraints. BIO constraints define valid BIO transitions for sequence tagging.}
\jy{For example, B-ARG0 cannot be followed by I-ARG1\cut{ and B-ARG0 cannot follow I-ARG0}.}
SRL constraints define rules on the role level and has three particular constraints:
unique core roles (U), continuation roles (C) and reference roles (R) \cite{punyakanok2008importance}. 
Lastly, syntactic constraints state \cut{the fact }that \textit{srl-spans}$(\y)$ have to be subset of 
\textit{parse-spans}$(\x)$.

\citep{he2017deep} use BIO and syntactic constraints at decoding time
by solving Eq.(\ref{eq:argmax_y}) where $f(x,y)$ incorporates those constraints and report that SRL constraints do not show significant improvements over the ensemble model. 
In particular, by using syntactic constraints, \citep{he2017deep} achieves up to $+2$ F1 score on CoNLL-$2005$ dataset via A* decoding.




Improvements of SRL system via use of syntactic constraints is consistent with other observations \cite{punyakanok2008importance}. However, all previous works enforce syntactic consistency only during decoding step. We propose that enforcing syntactic consistency during training time would also be beneficial and show the efficacy experimentally on Section \ref{sec:results}.

\paragraph{Enforcing Syntactic Consistency}
\label{sec:synt-const}


To quantify syntactic inconsistency, we define \textit{disagreeing-spans}$(\x, \y) = \{span_i \in \text{srl-spans}(\y)\ |\ span_i \notin \text{parse-spans}(\x)\}$. Further, we define \textit{disagreement rate}, $d(\x, \y)\in[0,1]$, and \textit{syntactic inconsistency score}, $s(\x, \y)\in [-1,1] $, as follows:
\begin{align}
d(\x, \y) = \frac{|\text{disagreeing-spans}(\x,\y)|}{|\text{srl-spans}(\y)|} \label{eq:disagree-rate} \\
s(\x, \y) =  2\times d(\x, \y) \label{eq:reward} -1 
\end{align} 

\paragraph{Syntactic Inconsistency Loss (SI-Loss)} For a given $(\x, v)$, let us consider $\hat{\y}^{(t)}$ to be the best possible tag sequence (according to Eq.(\ref{eq:argmax_y}) during epoch $t$ of model training. Ideally, if our model is syntax-aware, we would have $d(\x,\hat{\y}^{(t)}) = 0$ or $r(\x,\hat{\y}^{(t)}) = 1$. We define a loss component due to syntactic inconsistency (\textit{SI-Loss}) as follows:
\begin{align}
\text{SI-Loss} = s(\x,\hat{\y}^{(t)}) \sum_{i=1}^{|\hat{\y}^{(t)}|} \log p (\hat{y}^{(t)}_i| \x; \w^{(t)}) \label{eq:si-loss}
\end{align}

During training, we want to minimize SI-Loss.
 
\subsection{Training with Joint Objective}
\label{sec:joint-objective}
Based on Eq.(\ref{eq:nll-loss}), a supervised loss, and Eq.(\ref{eq:si-loss}), the SI-Loss, we propose a joint training objective. For a given sentence-predicate pair $(\x, v)$ and SRL tags $\y$, our joint training objective (at epoch t) is defined as:
\begin{align}
\text{Joint loss} = - \alpha_1 \sum_{i=1}^{|\y|} \log p(y_i|\x;\w) \nonumber \\
+ \alpha_2 \ r(\x,\hat{\y}^{(t)}) \sum_{i=1}^{|\hat{\y}^{(t)}|} \log p (\hat{y}^{(t)}_i| \x; \w^{(t)}) \label{eq:joint-obj}
\end{align}

Here, $\alpha_1$ and $\alpha_2$ are weights (hyper-parameters) for different loss components and are tuned using a development set. During training, we minimize joint loss - i.e., negative log-likelihood (or cross-entropy loss) and syntactic inconsistency loss.

\subsection{Semi-supervised learning formulation}
\label{sec:semi-sup}
In low resource scenarios, we have limited labeled data and larger amounts of unlabeled data. The obvious question is how to leverage large amounts of unlabeled data for training accurate models. In context of SRL, we propose to leverage SRL-unlabeled data in terms of parse trees. 

Observing Eq.(\ref{eq:si-loss}), one can notice that our formulation of SI-Loss is only dependent upon model's predicted tag sequence $\hat{\y}^{(t)}$ at a particular time point $t$ during training and the given sentence and it does not depend upon gold SRL tags. We leverage this fact in our SSL formulation to compute SI-loss from SRL-unlabeled sentences.

\cut{Towards this goal for SRL, we envision supervised data in terms of gold SRL tags and unsupervised data in terms of gold parse for sentences.}
Let \textit{sup-s} be a batch of SRL-labeled sentences 
and \textit{usup-s} be a batch SRL-unlabeled sentences only with parse information. In  SSL setup, we propose to train our model with joint objective where \textit{sup-s} only contributes to supervised loss Eq.(\ref{eq:nll-loss}) and \textit{unsup-s} contributes in terms of syntactic inconsistency objective Eq.(\ref{eq:si-loss}) and combine them according to Eq.(\ref{eq:joint-obj}) to train them with joint loss.
\tosvm{in terms of log-likelihood objective (as defined in Eq.(\ref{eq:nll-loss})) and \textit{unsup-s} contributes in terms of syntactic inconsistency objective (as defined in Eq(\ref{eq:si-loss})) 
and combine them according to Eq.(\ref{eq:joint-obj}) to train them with joint loss and back-propagate through it.}

%% file: 3.experiments.tex

\section{Experiments}
\label{sec:exps}

\begin{table}
	\small
	\centering
	\begin{tabular}{|c|r|r|}
		\hline
		Model/ & \multirow{2}{*}{Test F1} & Average\\
		Legend & & disagreement rate (\%)\\
		\hline
		B100 & 84.40 & 14.69 \\
		\textbf{B10} & $\mathbf{78.56}$ & $\mathbf{17.01}$ \\
		\textbf{B1}  & $\mathbf{67.28}$ & $\mathbf{21.17}$ \\
		\hline
		J100 &  84.75 $(+0.35)$ & 14.48 $(-1.43\%)$ \\
		J10 & 79.09 $(+0.53)$ & 16.25 $(-4.47\%)$\\
		J1 & 68.02 $(+0.74)$ & 20.49 $(-3.21\%)$\\
		\hline		
	\end{tabular}
	\caption{Comparison of baseline models (\textbf{B}) \cut{(trained without joint objective)} with the models trained with joint objective (\textbf{J}). \cut{Legend: \textbf{B$X$, J$X$} denotes model trained with $X$\% of the SRL-labeled data with respective objective.}} 
	\label{tab:joint-obj}
\end{table}

\subsection{Dataset}
\label{sec:dataset}
We evaluate our model's performance on span-based SRL dataset from CoNLL-2012 shared task  \cite{pradhan2013towards}. This dataset contains gold predicates as part of the input sentence and also gold parse information corresponding to each sentence which we use for defining hard constraints for SRL task. We use standard train/development/test split containing 278K/38.3K/25.6K sentences. \svm{Further, there is approx. 10\% disagreement between gold SRL-spans and gold parse-spans (we term these as noisy syntactic constraints). During training, we do not preprocess data to handle these noisy constraints but for the analysis related to enforcing syntactic constraints during inference, we study both cases: with and without noisy constraints. \footnote{\label{ft:noise}We preprocessed data for inference by simply deleting syntactic parse trees for the sentences where we have disagreement and perform standard Viterbi decoding for those sentences and note that this preprocessing scheme was never used during training.}}

\subsection{Model configurations}
\label{sec:model_configs}
For the SOTA system proposed in \cite{peters2018deep}, we use code from Allen AI\footnote{https://github.com/allenai/allennlp} to implement our approach. We follow their initialization and training configurations. \svm{ Let \textbf{B$\underline{X}$, J$\underline{X}$} denote model trained with $\underline{X}\%$ of the SRL-labeled data with cross-entropy and joint training objective, respectively.} \jy{B$\underline{X}$-SI$\underline{U}$x and B$\underline{X}$-J$\underline{U}$x denote model trained with SI-loss and Joint loss, respectively, on the pre-trained B$\underline{X}$ model where $\underline{X} \times \underline{U}$ amount of SRL-unlabeled data were used for further training.} 
To satisfy BIO constraints, we run Viterbi decoding by default for inference.

\begin{table}
	\small
	\centering
	\begin{tabular}{|c|r|r|}
		\hline
		Base Model/& \multirow{2}{*}{Test F1} & Average \\
		Legend & &disagreement rate (\%)\\
		\hline
		B10 & 78.56 & 17.01 \\ 
		B10-SI1x & 78.84 ($+0.28$) & 16.17 ($-4.94\%$) \\		
		B10-SI5x & 78.67 ($+0.11$) & 16.47 ($-3.17\%$) \\
		B10-SI10x & 78.76 ($+0.20$) & 16.09 ($-5.4\%$)\\
		\hline
		B1 &  67.28 & 21.17\\
		B1-SI1x &  67.67 ($+0.39$) & 20.14 ($-4.87\%$)\\		
		B1-SI5x &  67.74 ($+0.46$) & 19.93 ($-5.86\%$) \\
		B1-SI10x & 67.71 ($+0.43$) & 20.16 ($-4.77\%$) \\
		\hline		
	\end{tabular}
	\caption{Training with \textbf{SI-loss} for varying sizes of SRL-unlabeled data on top of the pre-trained baseline models 
	(B1, B10 on Table \ref{tab:joint-obj}). 
	}
	\label{tab:SI-only}
\end{table}

\subsection{Results}
We are interested in answering following questions. 
(Q1) how well does the baseline model produce syntactically consistent outputs, 
(Q2) does our approach actually enforce syntactic constraints,
(Q3) does our approach enforce syntactic constraints without compromising the quality of the system, 
(Q4) how well does our SSL formulation perform, especially in low-resource scenarios
, and lastly
(Q5) what is the difference in using the syntactic constraints in training time compared to using it at decoding time.
To answer (Q1-2) favorably we report average disagreement rate computed over test split. To answer (Q3-5), we report overall F1-scores on CoNLL-2012 test set (using standard evaluation script). \jy{For experiments using SRL-unlabeled data,} \svm{we report average results after running multiple experiments with different random samples of it.} 
\label{sec:results}

\begin{table}
	\small
	\centering
	\begin{tabular}{|c|r|r|r|}
		\hline
		Base Model/& \multirow{2}{*}{Test F1} & Average \\
		Legend & & disagreement rate (\%)\\
		\hline
		B10 & 78.56 & 17.01 \\ 
		B10-further & 78.86 ($+0.3$) & 16.25 ($-2.06\%$) \\
		B10-J1x & 79.23 ($+0.67$) & 16.03 ($-5.76\%$)\\		
		B10-J5x & 79.25 ($+0.69$) &  16.01 ($-5.88\%$)\\
		\textbf{B10-J10x} & \textbf{79.34 (+0.78)} & \textbf{15.88 (-6.64\%)}\\
		\hline
		B1 & 67.28 & 21.17\\ 
		B1-further & 67.76 ($+0.48$) & 20.75 ($-1.98\%$)\\
		B1-J1x & 68.45 ($+1.17$) & 19.57 ($-7.56\%$)\\		
		\textbf{B1-J5x} & \textbf{68.86 (+1.58)} & \textbf{19.38 (-8.46\%)}\\
		B1-J10x & 68.61 ($+1.33$) & 19.29 ($-8.88\%$)\\
		\hline		
	\end{tabular}
	\caption{Training with joint objective (\textbf{J}) on top of the baseline models (\textbf{B1}, \textbf{B10} from Table \ref{tab:joint-obj}), with the same SRL-labeled data used to train the baseline models and with varying sizes of SRL-unlabeled data. 
	}
	\label{tab:ssl-joint-obj}
\end{table}

\paragraph{Does training with joint objective help?}
We trained $3$ models with \svm{ random} 1\%, 10\% and \svm{ whole} 100\% of the training set with joint objective ($\alpha_1$ = $\alpha_2$ = $0.5$). For comparison, we trained $3$ SOTA models with the same training sets\cut{(i.e., $\alpha_1$ = $1.0$, $\alpha_2$ = $0.0$)}. All models were trained for max $150$ epochs and \cut{ended training earlier if validation F1 did not improve for}with a patience of $20$ epochs. Table \ref{tab:joint-obj} reports the results of this experiment. 
\svm{ We see that models trained with joint objective (J$X$) improve over baseline models (B$X$), both in terms of F1 and average disagreement rate.}
These improvements provide evidence for answering (Q1-3) favorably. Further, gains are more in low resource scenarios because by training models jointly to satisfy syntactic constraints helps in better generalization when trained with limited SRL corpora. 

\begin{table}
	\small
	\centering
	\begin{tabular}{|c|r|r|r|r|}
		\hline
		Decoding & B10 & B10-J10x & B1 & B1-J5x \\
		\hline
		Viterbi & 78.56 & 79.34 & 67.28 & 68.86\\
		\hline
		& \multicolumn{4}{c|}{Noisy syntactic constraints} \\ 
		\cline{2-5}
		\multirow{2}{*}{A*} & 72.95  & 73.57  & 63.77  & 64.73 \\
		& (-5.61) & (-5.77) & (-3.51) &  (-4.13) \\
		Gradient- & 79.7 & 80.21 & 69.85 & \textbf{70.95} \\
		based & (+1.41) & (+0.87) & (+2.57) & (\textbf{+2.1})\\
		\hline
		& \multicolumn{4}{c|}{Noise-free syntactic constraints} \\ 
		\cline{2-5}
		\multirow{2}{*}{A*} & 78.87  & 79.51  & 67.97  & 68.97 \\
		& (+0.31) & (+0.17) & (+0.69) &  (+0.11) \\
		Gradient- & 80.18 & \textbf{80.66} & 69.97 & 70.94 \\
		based  & (+1.62) & \textbf{(+1.32)} & (+2.69) & (+2.08)\\
		\hline	
	\end{tabular}
	\caption{Comparison of different decoding techniques: Viterbi, A* \cite{he2017deep} and gradient based inference \cite{lee2017enforcing} with noisy and noise-free syntactic constraints\footnotemark[1]. Note that the (+/-) F1 are reported w.r.t Viterbi decoding on the same column.}
	\label{tab:struct-predict}
\end{table}

\paragraph{Does SSL 
based training work for low-resource scenarios?}

\jy{To enforce syntactic constraints via SI-loss on SRL-unlabeled data,
we further train pre-trained model 
with two objectives in SSL set up:
(a) SI-loss (Table \ref{tab:SI-only}) and 
(b) joint objective (Table \ref{tab:ssl-joint-obj})} 

For experiment (a), we use square loss, $\beta \|W-W_{\text{pre-train}}\|^2$ regularizer to keep the model $W$ close to the pre-trained model $W_{\text{pre-train}}$ to avoid catastrophic forgetting ($\beta$ set to $0.005$). We optimize with SGD with learning rate of $0.01$, $\alpha_2 = 1.0$, patience of 10 epochs. 
\tosvm{Results related to this experiment are reported in Table \ref{tab:SI-only}.}\cut{We see improvements in range of 0.39-0.46 F1 and 0.11-0.28 F1 over B1 and B10, respectively. }\svm{We see that with SI-loss improvements are significant in terms of average disagreement rate as compared to F1.}

\svm{For experiment (b), we train B1 and B10 with joint objective in SSL set-up}\cut{wherein initial training set (SRL-labeled data) for B10 and B1 contribute in terms of log-likelihood objective and fine-tuning dataset (SRL-unlabeled data) contributes in terms of syntactic inconsistency objective} (as discussed in Section \ref{sec:semi-sup}). 
We use SGD with learning rate of $0.05$, $\alpha_1=\alpha_2=1.0$ and patience of 10 epochs. 
\tosvm{Table \ref{tab:ssl-joint-obj} report results for this experiment.} We report +1.58 F1 and +0.78 F1 improvement over B1 and B10, trained with 5\% and 100\% SRL-unlabeled data, respectively. \jy{Note that we cannot achieve these improvements with simply fine-tunning B$\underline{X}$ with supervised loss, as seen with B$\underline{X}$-further on Table \ref{tab:ssl-joint-obj}.}
 This provides evidence to answer (Q4) favorably. In general, the performance gains increase as the size of the SRL-unlabeled data increases. 

\paragraph{Is it better to enforce syntactic consistency on decoding or on training time?}

To answer (Q5), we conducted three experiments: using syntactic constraints on (a) inference only, i.e. structured prediction, (b) training only, and (c) both training and inference steps. For the structured prediction, we consider A* decoding, as used in \cite{he2017deep} and gradient-based inference \cite{lee2017enforcing}, which optimizes loss function similar to SI-loss on Eq.(\ref{eq:si-loss}) per example basis. If neither A* decoding nor gradient-based inference is used, we use Viterbi algorithm to enforce BIO constraints.
\tosvm{State the results for (a),(b),(c) }
The performance is the best (bold on Table \ref{tab:struct-predict}) when syntactic consistency is enforced both on training and inference steps, +3.67, +2.1 F1 score improvement over B1 and B10 respectively, and we conclude that the effort of enforcing syntactic consistency on inference time is complementary to the same effort on training time.
However, note that the overall performance increases as the benefit from enforcing syntactic consistency with SSL is far greater compared to marginal decrement on structured prediction.\\
While syntactic constraints help both train and inference, injecting constraints on train time is far more robust compared to enforcing them on decoding time. The performance of the structured prediction drops rapidly when the noise in the parse information is introduced ($x$ column of Table ~\ref{tab:struct-predict}). On the other hand, SSL was trained on CoNLL2012 data where about 10\% of the gold SRL-spans do not match with gold parse-spans and even when we increase noise level to 20\% the performance drop was only around 0.1 F1 score.

%% file: 4.related_work.tex

\section{Related Work}
\label{relatedWork}
The traditional approaches for SRL \cite{pradhan2005semantic, koomen2005generalized} constituted of cascaded system with four subtasks: pruning, argument identification, argument labeling, and inference. \cut{The pruning and argument identification steps used constituent parse trees and its relation to the given predicates. Then the subsequent model would learn to label provided arguments.}
\cut{The first effort towards end-to-end neural network architecture for SRL task was proposed by \cite{collobert2011natural}. They employed syntactic features from parse trees generated Charniak parser \cite{charniak2000maximum} and achieved better performance than any of the preexisting cascaded models.} Recent approaches \cite{zhou2015end, he2017deep} proposed end-to-end system for SRL using deep recurrent or bi-LSTM-based architecture with no syntactic inputs and have achieved SOTA results on English SRL. Lastly, \cite{peters2018deep} proposed ELMo, a deep contextualized word representations, and improved the SOTA English SRL by $3.2$ F1-points. 

\cut{Among the traditional four subtasks\cut{pruning, argument identification, argument labeling, and inference steps}}
Even on the end-to-end learning, inference still remains as a separate subtask and would be formalized as a constrained optimization problem. To solve this \svm{problem} ILP \cite{punyakanok2008importance}, {\em{A*} algorithm} \cite{he2017deep} and gradient-based inference \cite{lee2017enforcing} \svm{were employed. Further, all of these works leveraged syntactic parse during inference and was never used during training unless used as a cascaded system.}

To the best of our knowledge, this work is the first attempt towards SSL span-based SRL model. Nonetheless, there were few efforts in SSL in dependency-based SRL systems \cite{Furstenau2009-SSL-SRL,Deschacht2009-SSL-LM,Croce2010TowardOpen}. 
\cite{Furstenau2009-SSL-SRL} \svm{proposed to} augment the dataset by finding similar unlabeled sentences to already labeled set and annotate accordingly. \cut{\tosvm{ Let me know whether this makes sense.}}
While interesting, the similar augmentation technique is harder to apply to span-based SRL as one requires to annotate the whole span. \cite{Deschacht2009-SSL-LM, Croce2010TowardOpen} proposed to leverage the relation between words by learning latent word distribution over the context, i.e. language model. Our paper also incorporates this idea by using ELMo as it is trained via language model objective.

%% file: 5.conclusion.tex

\section{Conclusion and Future Work}

We presented a SI-loss to enforce SRL systems to produce syntactically consistent outputs. Further, leveraging the fact that SI-loss does not require labeled data, we proposed a SSL formulation with joint objective constituting of SI-loss and supervised loss together.
\jy{We show the efficacy of the proposed approach on low resource settings, +1.58, +0.78 F1 on 1\%, 10\% SRL-labeled data respectively, via further training on top of pre-trained SOTA model. We further show the structured prediction can be used as a complimentary tool and show performance gain of +3.67, +2.1 F1 over pre-trained model on 1\%, 10\% SRL-labeled data, respectively.
Semi-supervised training from the scratch and examination of semi-supervised setting on large dataset remains as part of the future work.}

%% file: appendix.tex
\section{Structured prediction formulation}
\citeauthor{he2017deep} proposed to incorporate such structural dependencies at decoding time by augmenting the loglikelihood function with penalization terms for constraint violations 
\begin{align}
\label{eq:score_function}
\f(\x, \y) = \sum_{i=1}^{n}\log p(y_i|\x) - \sum_{c \in C} c(\x, \y_{1:i})
\end{align}
where, each constraint function $c$ applies a non-negative penalty given the input $\x$ and a length-$t$ prefix $\y_{1:t}$.